# Principal Graphs and Manifolds

**Alexander N. Gorban**
*University of Leicester, United Kingdom*

**Andrei Y. Zinovyev**
*Institut Curie, Paris, France*

**ABSTRACT**

*In many physical, statistical, biological and other investigations it is desirable to approximate a system of points by objects of lower dimension and/or complexity. For this purpose, Karl Pearson invented principal component analysis in 1901 and found 'lines and planes of closest fit to system of points'. The famous k-means algorithm solves the approximation problem too, but by finite sets instead of lines and planes. This chapter gives a brief practical introduction into the methods of construction of general principal objects, i.e. objects embedded in the 'middle' of the multidimensional data set. As a basis, the unifying framework of mean squared distance approximation of finite datasets is selected. Principal graphs and manifolds are constructed as generalisations of principal components and k-means principal points. For this purpose, the family of expectation/maximisation algorithms with nearest generalisations is presented. Construction of principal graphs with controlled complexity is based on the graph grammar approach.*

INTRODUCTION

In many fields of science, one meets with multivariate (multidimensional) distributions of vectors representing some observations. These distributions are often difficult to analyse and make sense of due to the very nature of human brain which is able to visually manipulate only with the objects of dimension no more than three.

This makes actual the problem of approximating the multidimensional vector distributions by objects of lower dimension and/or complexity while retaining the most important information and structures contained in the initial full and complex data point cloud.

The most trivial and coarse approximation is collapsing the whole set of vectors into its *mean* point. The mean point represents the 'most typical' properties of the system, completely forgetting variability of observations.

The notion of the mean point can be generalized for approximating data by more complex types of objects. In 1901 Pearson proposed to approximate multivariate distributions by *lines* and *planes* (Pearson, 1901). In this way the Principal

Component Analysis (PCA) was invented, nowadays a basic statistical tool. Principal lines and planes go through the 'middle' of multivariate data distribution and correspond to the first few modes of the multivariate Gaussian distribution approximating the data.

Starting from 1950s (Steinhaus, 1956; Lloyd, 1957; and MacQueen, 1967), it was proposed to approximate the complex multidimensional dataset by several 'mean' points. Thus *k-means algorithm* was suggested and nowadays it is one of the most used *clustering methods* in machine learning (see a review presented by Xu & Wunsch, 2008).

Both these directions (PCA and K-Means) were further developed during last decades following two major directions: 1) linear manifolds were generalised for non-linear ones (in simple words, initial lines and planes were bended and twisted), and 2) some links between the 'mean' points were introduced. This led to appearance of several large families of new statistical methods; the most famous from them are Principal Curves, Principal Manifolds and Self-Organising Maps (SOM). It was quickly realized that the objects that are constructed by these methods are tightly connected theoretically. This observation allows now to develop a common framework called "Construction of Principal Objects". The geometrical nature of these objects can be very different but all of them serve as *data approximators of controllable complexity*. It allows using them in the tasks of *dimension* and *complexity reduction*. In Machine Learning this direction is connected with terms 'Unsupervised Learning' and 'Manifold Learning.'

In this chapter we will overview the major directions in the field of principal objects construction. We will formulate the problem and the classical approaches such as PCA and *k*-means in a unifying framework, and show how it is naturally generalised for the Principal Graphs and Manifolds and the most general types of principal objects, Principal Cubic Complexes. We will systematically introduce the most used ideas and algorithms developed in this field.

APPROXIMATIONS OF FINITE DATASETS

**Definition**. *Dataset* is a finite set $X$ of objects representing $N$ multivariate (multidimensional) observations. These objects $\mathbf{x}^i \in X$, $i = 1\ldots N$, are embedded in $\mathbf{R}^m$ and in the case of complete data are vectors $\mathbf{x}^i \in \mathbf{R}^m$. We will also refer to the individual components of $\mathbf{x}^i$ as $x_k^i$ such that $\mathbf{x}^i = (x_1^i, x_2^i, \ldots, x_m^i)$; we can also represent dataset as a *data matrix* $X = \{x_j^i\}$.

**Definition**. *Distance function* dist($\mathbf{x},\mathbf{y}$) is defined for any pair of objects $\mathbf{x}, \mathbf{y}$ from $X$ such that three usual axioms are satisfied: dist($\mathbf{x},\mathbf{x}$) = 0, dist($\mathbf{x},\mathbf{y}$) = dist($\mathbf{y},\mathbf{x}$), dist($\mathbf{x},\mathbf{y}$)+dist($\mathbf{y},\mathbf{z}$) $\leq$ dist($\mathbf{x},\mathbf{z}$).

**Definition**. *Mean point* $\mathbf{M}_F(X)$ for $X$ is a vector $\mathbf{M}_F \in \mathbf{R}^m$ such that $\mathbf{M}_F(X) = \arg\min_{\mathbf{y} \in R^m} \sum_{i=1..N} (\text{dist}(\mathbf{y},\mathbf{x}_i))^2$.

In this form the definition of the mean point goes back to Fréchet (1948). Notice that in this definition the mean point by Fréchet can be non-unique. However, this definition allows multiple useful generalisations including using it in the abstract metric spaces. It is easy to show that in the case of complete data and the Euclidean distance function $\text{dist}(\mathbf{x},\mathbf{y}) = \sqrt{\sum_{i=1}^{m}(\mathbf{x}^i - \mathbf{y}^i)^2}$, or, more generally, in the case of any quadratic distance function (for example, Mahalanobis distance), the mean point is the standard expectation $\mathbf{M}_F(X) = \frac{1}{N}\sum_{i=1}^{N}\mathbf{x}^i \equiv \mathbf{E}(X)$.

**Definition**. *Orthogonal projection* $P(\mathbf{x},Y)$ is defined for an object $\mathbf{x}$ and a set (not necessarily finite) of vectors $Y$ as a vector in $Y$ such that $P(\mathbf{x},Y) = \arg\min_{\mathbf{y}\in Y} \text{dist}(\mathbf{x},\mathbf{y})$. Notice that in principle, one can have non-unique and even infinitely many projections of $\mathbf{x}$ on $Y$.

**Definition**. *Mean squared distance* $\text{MSD}(X,Y)$ between a dataset $X$ and a set of vectors $Y$ is defined as $\text{MSD}(X,Y) = \sqrt{\frac{1}{N}\sum_{i=1}^{N}\text{dist}^2(\mathbf{x}^i,P(\mathbf{x}^i,Y))}$. We will also consider a simple generalisation of MSD: *weighted mean squared distance* $\text{MSD}_W(X,Y) = \sqrt{\frac{1}{\sum_{i=1}^{N}w_i}\cdot\sum_{i=1}^{N}w_i\,\text{dist}^2(\mathbf{x}^i,P(\mathbf{x}^i,Y))}$, where $w_i > 0$ is a weight for the object $\mathbf{x}_i$.

Our objective in the rest of the chapter is to briefly describe the methods for constructing various approximations (principal objects) for a dataset $X$. In almost all cases the principal objects will be represented as a finite or infinite set of vectors $Y \in \mathbf{R}^m$ such that 1) it approximates the finite dataset $X$ in the sense of minimisation of MSD($X,Y$), and 2) it answers some *regularity conditions* that will be discussed below.

PROBABILISTIC INTERPRETATION OF STATISTICS AND NOTION OF SELF-CONSISTENCY

In his original works, Pearson followed the principle that the only reality in data analysis is the dataset, embedded in a multidimensional metric space. This approach can be called *geometrical*. During the 20[th] century, probabilistic interpretation of statistics was actively developed. Accordingly to this interpretation, a dataset $X$ is one particular of i.i.d. sample from a multidimensional probability distribution $F(\mathbf{x})$ which defines a probability of appearance of a sample in the point $\mathbf{x} \in \mathbf{R}^m$.

The probability distribution, if can be estimated, provides a very useful auxiliary object allowing to define many notions in the theory of statistical data analysis. In particular, it allows us to define principal manifolds as self-consistent objects.

The notion of self-consistency in this context was first introduced by Efron (1967) and developed in the works of Flury (Tarpey & Flury, 1996), where it is claimed to be one of the most fundamental in statistical theory.

**Definition**. Given probability distribution $F(\mathbf{x})$ and a set of vectors $Y$ we say that $Y$ is *self-consistent with respect to $F(\mathbf{x})$* if $\mathbf{y} = \mathbf{E}_F(\mathbf{x}|P(\mathbf{x},Y) = \mathbf{y})$ for every vector $\mathbf{y} \in Y$. In words, it means that any vector $\mathbf{y} \in Y$ is a conditional mean expectation of point $\mathbf{x}$ under condition that $\mathbf{x}$ is orthogonally projected in $\mathbf{y}$.

The disadvantage of this definition for finite datasets is that it is not always possible to calculate the conditional mean, since typically for points $\mathbf{y} \in Y$ it is only one or zero point projected from $X$. This means that for finite datasets we should develop *coarse-grained self-consistency* notion. Usually it means that for every point $\mathbf{y} \in Y$ one defines some kind of neighbourhood and introduces a modified self-consistency with respect to this neighbourhood instead of $\mathbf{y}$ itself. Concrete implementations of this idea are described further in this chapter. In all cases, the effective size of the neighbourhood is a fundamental parameter in *controlling the complexity* of the resulting approximator $Y$.

FOUR APPROACHES TO CLASSICAL PCA

We can define linear principal manifolds as mean squared distance data approximators, constructed from linear manifolds embedded in $\mathbf{R}^m$. In fact, this corresponds to the original definition of principal lines and planes by Pearson (Pearson, 1901). However, PCA method was re-invented in other fields and even obtained different names (Karhunen-Loève or KL decomposition (Karhunen, 1946; Loève, 1955), Hotteling transform (Hotelling, 1933), Proper Orthogonal Decomposition (Lumley, 1967)) and others. Here we formulate four equivalent ways to define principal components that the user can meet in different applications.

Let us consider a linear manifold $L_k$ of dimension $k$ in the parametric form $L_k = \{\mathbf{a}_0 + \beta_1\mathbf{a}_1 + \ldots + \beta_k\mathbf{a}_k \mid \beta_i \in \mathbf{R}\}$, $\mathbf{a}_0 \in \mathbf{R}^m$ and $\{\mathbf{a}_1,\ldots, \mathbf{a}_k\}$ is a set of orthonormal vectors in $\mathbf{R}^m$.

**Definition** of PCA problem #1 (*data approximation by lines and planes*): PCA problem consists in finding such sequence $L_k$ ($k=1,2,\ldots,m-1$) that the sum of squared distances from data points to their orthogonal projections on $L_k$ is minimal over all linear manifolds of dimension $k$ embedded in $\mathbf{R}^m$:
$\text{MSD}(X, L_k) \to \min$  ($k=1,2,\ldots,m-1$).

**Definition** of PCA problem #2 (*variance maximisation*):
For a set of vectors $X$ and for a given $\mathbf{a}_i$, let us construct a one-dimensional distribution $\text{B}^i = \{\beta: \beta = (\mathbf{x},\mathbf{a}_i), \mathbf{x} \in X\}$ where $(\cdot,\cdot)$ denotes scalar vector product. Then let us define empirical variance of $X$ along $\mathbf{a}_i$ as $\text{Var}(\text{B}^i)$, where $\text{Var}(\ )$ is the standard empirical variance. PCA problem consists in finding such $L_k$ that the sum of empirical variances of $X$ along $\mathbf{a}_1,\ldots, \mathbf{a}_k$ would be maximal over all linear manifolds of dimension $k$ embedded in $\mathbf{R}^m$: $\sum_{i=1..k} \text{Var}(\text{B}^i) \to \max$. Let us also consider an orthogonal complement $\{\mathbf{a}_{k+1}, \ldots, \mathbf{a}_m\}$ of the basis $\{\mathbf{a}_1, \ldots, \mathbf{a}_k\}$. Then an equivalent definition (*minimization of residue variance*) is

$$\sum_{i=k+1}^{m} \mathrm{Var}(\mathrm{B}^i) \to \min .$$

**Definition** of PCA problem #3 (*mean point-to-point squared distance maximisation*):
PCA problem consists in finding such sequence $L_k$ that the mean point-to-point squared distance between the orthogonal projections of data points on $L_k$ is maximal over all linear manifolds of dimension $k$ embedded in $\mathbf{R}^m$:

$$\frac{1}{N}\sum_{i,j=1}^{N} \mathrm{dist}^2(P(\mathbf{x}^i, L_k), P(\mathbf{x}^j, L_k)) \to \max .$$

Having in mind that all orthogonal projections onto lower-dimensional space lead to contraction of all point-to-point distances (except for some that do not change), this is equivalent to minimisation of *mean squared distance distortion*:

$$\sum_{i,j=1}^{N} [\mathrm{dist}^2(\mathbf{x}^i, \mathbf{x}^j) - \mathrm{dist}^2(P(\mathbf{x}^i, L_k), P(\mathbf{x}^j, L_k))] \to \min .$$

In the three above mentioned definitions, the basis vectors are defined up to an arbitrary rotation that does not change the manifold. To make the choice less ambiguous, in the PCA method the following principle is applied: given $\{a_0, a_1, \ldots, a_k\}$, any 'embedded' linear manifold of smaller dimension $s$ in the form $L_s = \{a_0 + \beta_1 a_1 + \ldots + \beta_s a_s | \beta_i \in \mathbf{R}, s < k\}$, must be itself a linear principal manifold of dimension $s$ for $X$ (a *flag* of principal subspaces).

**Definition** of PCA problem #4 (*correlation cancellation*):
Find such an orthonormal basis $(\mathbf{a}_1, \ldots, \mathbf{a}_s)$ in which the covariance matrix for $\mathbf{x}$ is diagonal. Evidently, in this basis the distributions $(\mathbf{a}_i, \mathbf{x})$ and $(\mathbf{a}_j, \mathbf{x})$, for $i \neq j$, have zero correlation.

Definitions 1-3 were given for finite datasets while definition 4 is sensible both for finite datasets and random vector $\mathbf{x}$. For finite datasets the empiric correlation should be cancelled. The empiric principal components which annul empiric correlations could be considered as an approximation to the principal components of the random vector.

Equivalence of the above-mentioned definitions in the case of complete data and Euclidean space follows from Pythagorean Theorem and elementary algebra. However, in practice this or that definition can be more useful for computations or generalisations of the PCA approach. Thus, only definitions #1 and #3 are suitable for working with incomplete data since they are defined with use of only distance function that can be easily calculated for the 'gapped' data vectors (see further). The definition #1 can be generalized by weighting data points (Cochran & Horne, 1977), while the definition #3 can be generalized by weighting pairs of data points (Gabriel & Zamir, 1979). More details about PCA and generalisations could be found in the fundamental book by Jollliffe (2002).

BASIC EXPECTATION/MAXIMISATION ITERATIVE ALGORITHM FOR FINDING PRINCIPAL OBJECTS

Most of the algorithms for finding principal objects for a given dataset *X* are constructed accordingly to the classical expectation/maximisation (EM) splitting scheme that was first formulated as a generic method by Dempster et al (1977):

Generic Expectation-Maximisation algorithm for estimating principal objects

1) *Initialisation step*. Some initial configuration of the principal object *Y* is generated;
2) *Expectation (projection) step*. Given configuration of *Y*, calculate orthogonal projections $P(\mathbf{x},Y)$, for all $\mathbf{x} \in X$;
3) *Maximisation step*. Given the calculated projections, find more optimal configuration of *Y* with respect to *X*.
4) *(Optional) adaptation step*. Using some strategy, change the properties of *Y* (typically, add or remove points to *Y*).
5) *Repeat steps 2-4 until some convergence criteria would be satisfied.*

For example, for the principal line, we have the following implementation of the above mentioned bi-iteration scheme (Bauer, 1957; for generalisations see works of Roweis (1998) and Gorban & Rossiev (1999)).

Iterative algorithm for calculating the first principal component

1) Set $\mathbf{a}_0 = \mathbf{M}_F(X)$ (i.e., zero order principal component is the mean point of *X*);
2) Choose randomly $\mathbf{a}_1$;
3) Calculate $b_i = \dfrac{(\mathbf{x}_i - \mathbf{a}_0, \mathbf{a}_1)}{\|\mathbf{a}_1\|^2}$, $i = 1\ldots N$;
4) Given $b_i$, find new $\mathbf{a}_1$, such that $\sum_{i=1}^{N}(\mathbf{x}_i - \mathbf{a}_0 - \mathbf{a}_1 b_i)^2 \to \min_{\mathbf{a}_1}$, i.e.
$$\mathbf{a}_1 = \frac{\sum_{i=1..N}\mathbf{x}_i b_i - \mathbf{a}_0 \sum_{i=1..N} b_i}{\sum_{i=1..N} b_i^2};$$
5) Re-normalize $\mathbf{a}_1 := \mathbf{a}_1 / \|\mathbf{a}_1\|$.
6) Repeat steps 3-5 until the direction of $\mathbf{a}_1$ do not change more than on some small angle ε.

**Remark.** To calculate all other principal components, *deflation* approach is applied: after finding $\mathbf{a}_1$, one calculates new $X^{(1)} = X - \mathbf{a}_0 - \mathbf{a}_1(\mathbf{x},\mathbf{a}_1)$, and the procedure is repeated for $X^{(1)}$.

**Remark.** The basic EM procedure has good convergence properties only if the first eigenvalues of the empirical covariance matrix $X^T X$ are sufficiently well separated. If this is not the case, more sophisticated approaches are needed (Bau & Trefethen, 1997).

The PCA method can be treated as *spectral decomposition* of the symmetric and positively defined empirical covariance data matrix (defined in the case of complete data) $C = \frac{1}{N-1} X^T X$ or $C_{ij} = \frac{1}{N-1} \sum_{k=1}^{N} x_i^k x_j^k$, where without loss of generality we suppose that the data are centered.

**Definition.** We call $\sigma > 0$ a *singular value* for the data matrix $X$ iff there exist two vectors of unit length $\mathbf{a}_\sigma$ and $\mathbf{b}_\sigma$ such that $X\mathbf{a}_\sigma = \sigma \mathbf{b}_\sigma^T$ and $\mathbf{b}_\sigma X = \sigma \mathbf{a}_\sigma^T$. Then the vectors $\mathbf{a}_\sigma = \{ a_1^{(\sigma)}, \cdots, a_m^{(\sigma)} \}$ and $\mathbf{b}_\sigma = \{ b_1^{(\sigma)}, \cdots, b_N^{(\sigma)} \}$ are called *left and right singular vectors for the singular value* $\sigma$.

If we know all $p$ singular values of $X$, where $p = \text{rank}(X) \leq \min(N, m)$, then we can represent $X$ as $X = \sum_{l=1}^{p} \sigma_l \mathbf{b}_{(l)} \mathbf{a}_{(l)}$ or $x_i^k = \sum_{l=1}^{p} \sigma_l b_k^{(l)} a_i^{(l)}$. It is called the *singular value decomposition* (SVD) of $X$. It is easy to check that the vectors $\mathbf{a}_{(l)}$ correspond to the principal vectors of $X$ and the eigenvectors of the empirical covariance matrix $C$, whereas $\mathbf{b}_{(l)}$ contain projections of $N$ points onto the corresponding principal vector. Eigenvalues $\lambda_l$ of $C$ and singular values $\sigma_l$ of $X$ and are connected by $\lambda_l = \frac{1}{N-1} (\sigma_l)^2$.

The mathematical basis for SVD was introduced by Sylvester (1889) and it represents a solid mathematical foundation for PCA (Strang, 1993). Although formally the problems of spectral decomposition of $X$ and eigen decomposition of $C$ are equivalent, the algorithms for performing singular decomposition directly (without explicit calculation of $C$) can be more efficient and robust (Bau III & Trefethen, 1997). Thus, the iterative EM algorithm for calculating the first principal component described in the previous chapter indeed performs singular decomposition (for centered data we simply put $\mathbf{a}_0 = \mathbf{0}$) and finds right singular (principal) and left singular vectors one by one.

K-MEANS AND PRINCIPAL POINTS

*K-means* clustering goes back to 1950s (Steinhaus (1956); Lloyd (1957); and MacQueen (1967)). It is another extreme in its simplicity case of finding a principal object. In this case it is simply an unstructured finite (and usually, much smaller than the number of points $N$ in the dataset $X$) set of vectors (centroids). One can say that the solution searched by the *k*-means algorithm is a set of $k$ principal points (Flury, 1990).

**Definition.** A set of $k$ points $Y=\{\mathbf{y}_1,..,\mathbf{y}_k\}$, $\mathbf{y}_i \in \mathbf{R}^m$ is called *a set of principal points* for dataset $X$ if it approximates $X$ with minimal mean squared distance error over all sets of *k*-points in $\mathbf{R}^m$ (distortion): $\sum_{\mathbf{x} \in X} \text{dist}^2(\mathbf{x}, P(\mathbf{x}, Y)) \to \min$, where $P(\mathbf{x}, Y)$ is the point from $Y$ closest to $\mathbf{x}$. Note that the set of principal points can be not unique.

The simplest implementation of the *k*-means procedure follows the classical EM scheme:

Basic *k*-means algorithm

1) Choose initial position of $\mathbf{y}_1,...,\mathbf{y}_k$ randomly from $\mathbf{x}_i \in X$ (with equal probabilities);
2) Partition $X$ into subsets $K_i$, $i=1..k$ of data points by their proximity to $\mathbf{y}_k$:
$K_i = \{\mathbf{x} : \mathbf{y}_i = \arg\min_{\mathbf{y}_j \in Y} \text{dist}(\mathbf{x}, \mathbf{y}_j)\}$;
3) Re-estimate $\mathbf{y}_i = \frac{1}{|K_i|} \sum_{\mathbf{x} \in K_i} \mathbf{x}$, $i = 1..k$;
4) Repeat steps 2-3 until complete convergence.

The method is sensitive to the initial choice of $\mathbf{y}_1,...,\mathbf{y}_k$. Arthur & Vassilvitskii (2007) demonstrated that the special construction of probabilities instead of equidistribution gives serious advantages. The first centre, $\mathbf{y}_1$, they select equiprobable from *X*. Let the centres $\mathbf{y}_1,...,\mathbf{y}_j$ are chosen ($j < k$) and $D(\mathbf{x})$ be the squared shortest distance from a data point $\mathbf{x}$ to the closest centre we have already chosen. Then, we select the next centre, $\mathbf{y}_{j+1}$, from $\mathbf{x}_i \in X$ with probability
$$p(\mathbf{x}_i) = D(\mathbf{x}_i) \Big/ \sum_{\mathbf{x} \in X} D(\mathbf{x}).$$

Evidently, any solution of *k*-means procedure converges to a self-consistent set of points $Y = \{\mathbf{y}_1,...,\mathbf{y}_k\}$ (because $Y = E[P(X,Y)]$), but this solution may give a local minimum of distortion and is not necessary the set of principal points (which is the globally optimal approximator from all possible *k*-means solutions).

Multiple generalisations of *k*-means scheme have been developed (see, for example, a book of Mirkin (2005) based on the idea of 'data recovering'). The most computationally expensive step of the algorithm, partitioning the dataset by proximity to the centroids, can be significantly accelerated using *kd*-tree data structure (Pelleg & Moore, 1999). Analysis of the effectiveness of EM algorithm for the *k*-means problem was given by Ostrovsky et al. (2006).

Notice that the case of principal points is the only in this chapter when self-consistency and coarse-grained self-consistency coincide: centroid $\mathbf{y}_k$ is the conditional mean point for the data points belonging to the Voronoi region associated with $\mathbf{y}_k$.

LOCAL PCA

The term 'Local PCA' was first used by Braverman (1970) and Fukunaga & Olsen (1971) to denote the simplest cluster-wise PCA approach which consists in 1) applying *k*-means or other type of clustering to a dataset and 2) calculating the principal components for each cluster separately. However, this simple idea performs rather poorly in applications, and more interesting approach consists in generalizing *k*-means by introducing *principal hyperplane segments* proposed by Diday (1979) and called '*k*-segments' or *local subspace analysis* in a more advanced version (Liu, 2003). The algorithm for their estimation follows the classical EM scheme.

Further development of the local PCA idea went in two main directions. First, Verbeek (2002) proposed a variant of the '*k*-segment' approach for one-dimensional segments accompanied by a strategy to assemble disconnected line segments into the global piecewise linear principal curve. Einbeck et al (2008) proposed an iterative cluster splitting and joining approach (*recursive local PCA*) which helps to select the optimal number and configuration of disjoined segments.

Second direction is associated with a different understanding of 'locality'. It consists in calculating *local mean points* and *local principal directions* and following them starting from (may be multiple) seed points. Locality is introduced using kernel functions defining the effective radius of neighborhood in the data space. Thus, Delicado (2001) introduced *principal oriented points* (POP) based on the variance maximisation-based definition of PCA (#2 in our chapter). POPs are different from the principal points introduced above because they are defined independently one from another, while the principal points are defined globally, as a set. POPs can be assembled into the *principal curves of oriented points* (PCOP). Einbeck (2005) proposed a simpler approach based on local tracing of principal curves by calculating *local centers of mass* and the *local first principal components*.

SOM APPROACH FOR PRINCIPAL MANIFOLD APPROXIMATION AND ITS GENERALISATIONS

Kohonen in his seminal paper (Kohonen, 1982) proposed to modify the *k*-means approach by introducing connections between centroids such that a change in the position of one centroid would also change the configuration of some neighboring centroids. Thus Self-Organizing Maps (SOM) algorithm was developed.

With the SOM algorithm (Kohonen, 1982) we take a finite metric space *V* with metric $\rho$ and try to map it into $\mathbf{R}^m$ with combinations of two criteria: (1) the best preservation of initial structure in the image of *V* and (2) the best approximation of the dataset *X*. In this way, SOMs give the most popular approximations for principal manifolds: we can take for *V* a fragment of a regular *s*-dimensional grid and consider the resulting SOM as the approximation to the *s*-dimensional principal manifold (Mulier & Cherkassky, 1995; Ritter et al, 1992; Yin H. 2008).

The SOM algorithm has several setup variables to regulate the compromise between these goals. In the original formulation by Kohonen, we start from some initial approximation of the map, $\phi_1: V \to \mathbf{R}^m$. Usually this approximation lies on the *s*-dimensional linear principal manifold. On each *k*-th step of the algorithm we have a chosen datapoint $\mathbf{x} \in X$ and a current approximation $\phi_k: V \to \mathbf{R}^m$. For these $\mathbf{x}$ and $\phi_k$ we define an 'owner' of $\mathbf{x}$ in *V*: $v_x = \arg\min_{v \in V} \|\mathbf{x} - \phi_k(v)\|$. The next approximation, $\phi_{k+1}$, is $\phi_{k+1}(v) = h_k \times w(\rho(v,v_x))(\mathbf{x} - \phi_k(v))$. Here $h_k$ is a step size, $0 \leq w(\rho(v,v_x)) \leq 1$ is a monotonically decreasing neighborhood function. This process proceeds in several epochs, with neighborhood radius decreasing during each next epoch.

The idea of SOM is flexible, was applied in many domains of science, and it lead to multiple generalizations (see the review paper by Yin (2008)). Some of the algorithms for constructing SOMs are of EM type described above, such as the *Batch SOM Algorithm* (Kohonen, 1997): it includes projecting step exactly the same as in *k*-means and the maximization step at which all $\phi_k(v)$ are modified simultaneously.

One source of theoretical dissatisfaction with SOM is that it is not possible to define an optimality criterion (Erwin et al, 1992): SOM is a result of the algorithm at work and there does not exist any objective function that is minimized by the training process.

In attempt to resolve this issue, Bishop et al. (1998) developed the optimization-based Generative Topographic Mapping (GTM) method. In this setting, it is supposed that the observed data is i.i.d. sample from a mixture of Gaussian distributions with the centers aligned along a two-dimensional grid, embedded in the data space. Parameters of this mixture are determined by EM-based maximization of the likelihood function (probability of observing *X* within this data model).

PRINCIPAL MANIFOLDS BY HASTIE AND STUELZE

Principal curves and principal two-dimensional surfaces for a probability distribution $F(\mathbf{x})$ were introduced in the PhD thesis by Trevor Hastie (1984) as a self-consistent (non-linear) one- and two-dimensional globally parametrisable smooth manifolds without self-intersections.

**Definition**. Let *G* be the class of differentiable 1-dimensional curves in $\mathbf{R}^m$, parameterized by $\lambda \in \mathbf{R}^1$ and without self-intersections. The *Principal Curve* of the probability distribution $F(\mathbf{x})$ is such a $Y(\lambda) \in G$ that is self-consistent.

**Remark.** Usually, a compact subset of $\mathbf{R}^m$ and a compact interval of parameters $\lambda \in \mathbf{R}^1$ are considered. To discuss unbounded regions, it is necessary to add a condition that $Y(\lambda)$ has finite length inside any bounded subset of $\mathbf{R}^m$ (Kégl, 1999).

**Definition**. Let $G^2$ be the class of differentiable 2-dimensional surfaces in $\mathbf{R}^m$, parameterized by $\lambda \in \mathbf{R}^2$ and without self-intersections. The *Principal Surface* of the probability distribution $F(\mathbf{x})$ is such a $Y(\lambda) \in G^2$ that is self-consistent. (Again, for unbounded regions it is necessary to assume that for any bounded set *B* from $\mathbf{R}^m$ the set of parameters $\lambda$ for which $Y(\lambda) \in B$ is also bounded.)

First, Hastie and Stuelze proposed an algorithm for finding the principal curves and principal surfaces for a probability distribution $F(\mathbf{x})$, using the classical EM splitting. We do not provide this algorithm here because for a finite dataset *X* it can not be directly applied because in a typical point on $Y(\lambda)$ only zero or one data point is projected, hence, one can not calculate the expectation. As mentioned above, in this case we should use some kind of coarse-grained self-consistency. In the original approach by Hastie (1984), this is done through introducing

*smoothers*. This gives the practical formulation of the HS algorithm for estimating the principal manifolds from a finite dataset *X*:

Hastie and Stuelze algorithm for finding principal curve for finite dataset

1) Initialize $Y(\lambda) = \mathbf{a}_0 + \lambda \mathbf{a}_1$, where $\mathbf{a}_0$ is a mean point and $\mathbf{a}_1$ is the first principal component;
2) Project every data point $\mathbf{x}_i$ onto $Y(\lambda)$: i.e., for each $\mathbf{x}_i$ find $\lambda_i$ such that $Y(\lambda_i) = \arg\inf_{\lambda} \| Y(\lambda) - \mathbf{x}_i \|^2$. In practice it requires interpolation procedure because $Y(\lambda)$ is determined in a finite number of points $\{\lambda_1,...,\lambda_N\}$. The simplest is the piecewise interpolation procedure, but more sophisticated procedures can be proposed (Hastie, 1984);
3) Calculate new $Y'(\lambda)$ in the finite number of internal coordinates $\{\lambda_1,...,\lambda_N\}$ (found at the previous step) as the local average of points $\mathbf{x}_i$ and some other points, that have close to $\lambda_i$ projections onto $Y$. To do this, 1) a *span* [$w \times N$] is defined ( [.] here is integer part ), where $0 < w \ll 1$ is a parameter of the method (coarse-grained self-consistency neighbourhood radius); 2) for [$w \times N$] internal coordinates $\{\lambda_{i_1},...,\lambda_{i_{[w \times N]}}\}$ closest to $\lambda_i$ and the corresponding $\{\mathbf{x}_{i_1},...,\mathbf{x}_{i_{[w \times N]}}\}$ calculate weighted least squares linear regression $\mathbf{y}(\lambda) = \mathbf{a}^{(i)}\lambda + \mathbf{b}^{(i)}$; 3) define $Y'(\lambda_i)$ as the value of the linear regression in $\lambda_i$: $Y'(\lambda_i) = \mathbf{a}^{(i)}\lambda_i + \mathbf{b}^{(i)}$.
4) Reassign $Y(\lambda) \leftarrow Y'(\lambda)$
5) Repeat steps 2)-4) until $Y$ does not change (approximately).

**Remark.** For the weights in the regression at the step 3) Hastie proposed to use some symmetric kernel function that vanishes on the borders of the neighbourhood. For example, for $\mathbf{x}_i$ let us denote as $\lambda_{i_{[w \times N]}}$ the most distant value of the internal coordinate from [$w \times N$] ones closest to $\lambda_i$. Then we can define weight for the pair $(\lambda_{i_j}, \mathbf{x}_{i_j})$ as

$$\omega_j^i = \begin{cases} (1 - (|\lambda_{i_j} - \lambda_i|/|\lambda_{i_j} - \lambda_{i_N}|)^3)^{1/3}, & if \ |\lambda_{i_j} - \lambda_i| < |\lambda_{i_j} - \lambda_{i_N}|, \\ 0, & otherwise. \end{cases}$$

**Remark.** At the step 3) an alternative approach was also proposed with use of cubic splines to approximate the smooth function $Y'(\lambda)$ from all pairs $(\lambda_i, \mathbf{x}_i)$, $i = 1..N$.

Non-linear Principal Manifolds constructed by this algorithm are usually called *Hastie-Stuelze* (*HS*) *principal manifolds*. However, the global optimality of HS principal manifolds is not guaranteed (only self-consistency in the case of distribution or coarse-grained self-consistency in the case of dataset is guaranteed by construction). For example, the second principal component of a sample *X* from a normal distribution is self-consistent and will be correct HS principal curve but of course not the optimal one.

We should also underline that our view on what is the object constructed by the HS algorithm for a dataset *X* depends on 1) probabilistic interpretation of the nature of *X*, and 2) the chosen heuristic approach to coarse-grained self-consistency. If we do not suppose that the dataset is generated by i.i.d. sampling from $F(\mathbf{x})$ then the definition of HS principal manifold is purely operational: HS principal manifold for $X$ is the result of application of HS algorithm for finite datasets. Analogous remark is applicable for all principal manifold approximators constructed for finite datasets and described further in this chapter.

In his PhD thesis Hastie noticed that the HS principal curve does not coincide with the generating curve in a very simple additive data generation model

$$X = f(\lambda) + \varepsilon, \qquad (1)$$

where $f(\lambda)$ is some curve embedded in data space and $\varepsilon$ is noise distribution independent on $\lambda$. Because of the fact that if $f(\lambda)$ is not a straight line then it is not self-consistent, HS principal curves were claimed to be 'biased'. This inspired Tibshirani (1992) to introduce an alternative definition of the principal curve, based directly on a continuous mixture model (1) and maximising regularized likelihood.

KÉGL-KRYZHAK IMPROVEMENT

Kégl in his PhD thesis supervised by Kryzhak (Kégl, 1999) revised the existing methods for estimating the principal curves. In particular, this led to the definition of principal curves with limited length.

**Definition**. *Principal curve $Y_L(\lambda)$ of length $L$ is such a curve that the mean squared distance from the dataset $X$ to the curve $Y_L(\lambda)$ is minimal over all curves of length less than or equal to $L$*: $\sum_{i=1}^{N} \text{dist}^2(\mathbf{x}^i, P(\mathbf{x}^i, Y_L)) \to \min$.

**Theorem.** Assume that $X$ has finite second moments, i.e. $\sum_{i=1}^{N} \mathbf{x}^i (\mathbf{x}^i)^T < \infty$. Then for any $L > 0$ there exists a principal curve of length $L$.

Principal curves of length $L$ as defined by Kégl, are globally optimal approximators as opposite to the HS principal curves that are only self-consistent. However, all attempts to construct a practical algorithm for finding globally optimal principal curves of length $L$ were not successful. Instead Kégl developed an efficient heuristic *Polygonal line algorithm* for constructing piecewise linear principal curves.

Let us consider a piecewise curve $Y$ composed from vertices located in points $\{\mathbf{y}^1,\ldots,\mathbf{y}^{k+1}\}$ and $k$ segments connecting pairs of vertices $\{\mathbf{y}^j,\mathbf{y}^{j+1}\}$, $j=1..k$. Kégl's algorithm searches for a (local) optimum of the penalised mean squared distance error function:

$$U(X,Y) = \text{MSD}(X,Y) + \frac{\lambda}{k+1}\sum_{i=1}^{k+1}\text{CP}(i), \qquad (2)$$

where $\text{CP}(i)$ is a curvature penalty function for a vertex $i$ chosen as

$$CP(i) = \begin{cases} \|\mathbf{y}^1 - \mathbf{y}^2\|^2 & \text{if } i = 1 \\ r^2(1 + \cos\gamma(i)) & \text{if } 1 < i < k+1 \\ \|\mathbf{y}^k - \mathbf{y}^{k+1}\|^2 & \text{if } i = k+1 \end{cases},$$

where $\cos\gamma(i) = \dfrac{(\mathbf{y}^{i-1} - \mathbf{y}^i, \mathbf{y}^{i+1} - \mathbf{y}^i)}{\|\mathbf{y}^{i-1} - \mathbf{y}^i\| \|\mathbf{y}^{i+1} - \mathbf{y}^i\|}$ is the cosines of the angle between two neighbouring segments at the vertex $i$, $r = \max_{\mathbf{x} \in X} \text{dist}(\mathbf{x}, \mathbf{M_F}(X))$ is the 'radius' of the dataset $X$, and $\lambda$ is a parameter controlling the curve global smoothness.

The *Polygonal line algorithm* (Kégl, 1999) follows the standard EM splitting scheme:

Polygonal line algorithm for estimating piece-wise linear principal curve

1) The initial approximation is constructed as a segment of principal line. The length of the segment is the difference between the maximal and the minimal projection value of $X$ onto the first principal component. The segment is positioned such that it contains all of the projected data points. Thus in the initial approximation one has two vertices $\{\mathbf{y}^1, \mathbf{y}^2\}$ and one segment between them ($k = 1$).
2) *Projection step.* The dataset $X$ is partitioned into $2k+1$ $K_z = \{\mathbf{x} : z = \arg\min_{z \in \text{vertices} \cup \text{segments}} \text{dist}(\mathbf{x}, z)\}$ subsets constructed by their proximity to $k+1$ vertices and $k$ segments. If a segment $i$ and a vertex $j$ are equally distant from $\mathbf{x}$ then $\mathbf{x}$ is placed into $K_j$ only.
3) *Optimisation step.* Given partitioning obtained at the step 2, the functional $U(X,Y)$ is optimised by use of a gradient technique. Fixing partitioning into $K_i$ is needed to calculate the gradient of $U(X,Y)$ because otherwise it is not a differentiable function with respect to the position of vertices $\{\mathbf{y}_i\}$.
4) *Adaptation step.* Choose the segment with the largest number of points projected onto it. If more than one such segment exists then the longest one is chosen. The new vertex is inserted in the midpoint of this segment; all other segments are renumerated accordingly.
5) *Stopping criterion.* The algorithm stops when the number of segments exceeds $\beta \cdot N^{1/3} \cdot \dfrac{r}{\text{MSD}(X,Y)}$.

Heuristically, the default parameters of the method have been proposed $\beta = 0.3$, $\lambda = \lambda' \cdot \dfrac{k}{N^{1/3}} \cdot \dfrac{\text{MSD}(X,Y)}{r}$, $\lambda' = 0.13$. The details of implementation together with convergence and computational complexity study are provided elsewhere (Kégl, 1999).

Smola et al. (2001) proposed a *regularized principal manifolds* framework, based on minimization of quantization error functional with a large class of regularizers that can be used and a universal EM-type algorithm. For this algorithm, the convergence rates were analyzed and it was showed that for some regularizing terms the convergence can be optimized with respect to the Kegl's polygonal line algorithm.

ELASTIC MAPS APPROACH

In a series of works (Gorban & Rossiev, 1999; Gorban et al., 2001, 2003; Gorban & Zinovyev, 2005, 2008a; Gorban et al., 2007, 2008), the authors of this chapter used metaphor of elastic membrane and plate to construct one-, two- and three-dimensional principal manifold approximations of various topologies. Mean squared distance approximation error combined with the elastic energy of the membrane serves as a functional to be optimised. The elastic map algorithm is extremely fast at the optimisation step due to the simplest form of the smoothness penalty. It is implemented in several programming languages as software libraries or front-end user graphical interfaces freely available from the web-site http://bioinfo.curie.fr/projects/vidaexpert. The software found applications in microarray data analysis, visualization of genetic texts, visualization of economical and sociological data and other fields (Gorban et al, 2001, 2003; Gorban & Zinovyev 2005, 2008a; Gorban et al, 2007, 2008).

Let $G$ be a simple undirected graph with set of vertices $V$ and set of edges $E$.

**Definition**. *k-star* in a graph $G$ is a subgraph with $k + 1$ vertices $v_{0,1,...,k} \in V$ and $k$ edges $\{(v_0, v_i)/i = 1, ..., k\} \in E$. The *rib* is by definition a 2-star.

**Definition**. Suppose that for each $k \geq 2$, a family $S_k$ of $k$-stars in $G$ has been selected. Then we define an *elastic graph* as a graph with selected families of $k$-stars $S_k$ and for which for all $E^{(i)} \in E$ and $S_k^{(j)} \in S_k$, the corresponding elasticity moduli $\lambda_i > 0$ and $\mu_{kj} > 0$ are defined.

**Definition**. *Primitive elastic graph* is an elastic graph in which every non-terminal node (with the number of neighbours more than one) is associated with a $k$-star formed by *all* neighbours of the node. All $k$-stars in the primitive elastic graph are selected, i.e. the $S_k$ sets are completely determined by the graph structure.

**Definition**. Let $E^{(i)}(0)$, $E^{(i)}(1)$ denote two vertices of the graph edge $E^{(i)}$ and $S_k^{(j)}(0)$, ..., $S_k^{(j)}(k)$ denote vertices of a $k$-star $S_k^{(j)}$ (where $S_k^{(j)}(0)$ is the central vertex, to which all other vertices are connected). Let us consider a map $\phi: V \to \mathbf{R}^m$ which describes an embedding of the graph into a multidimensional space. The *elastic energy of the graph embedding in the Euclidean space* is defined as

$$U^\phi(G) := U_E^\phi(G) + U_R^\phi(G), \tag{3}$$

$$U_E^\phi(G) := \sum_{E^{(i)}} \lambda_i \left\| \phi(E^{(i)}(0)) - \phi(E^{(i)}(1)) \right\|^2, \tag{4}$$

$$U_E^\phi(G) := \sum_{S_k^{(j)}} \mu_{kj} \| \phi(S_k^{(j)}(0)) - \frac{1}{k} \sum_{i=1}^{k} \phi(S_k^{(j)}(i)) \|^2. \tag{5}$$

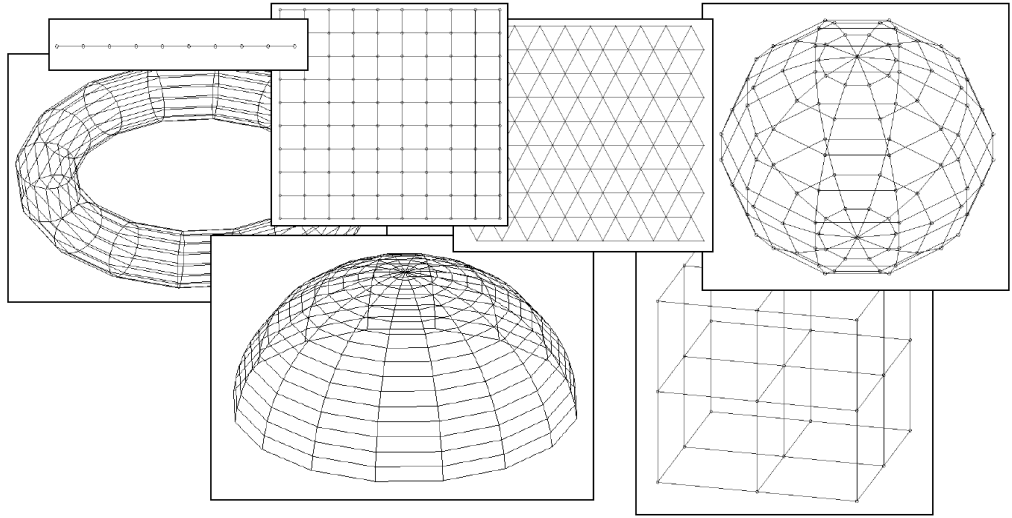

*Fig. 1. Elastic nets used in practice.*

**Definition**. *Elastic net* is a particular case of elastic graph which (1) contains only ribs (2-stars) (the family $S_k$ are empty for all k>2); and (2) the vertices of this graph form a regular small-dimensional grid (Fig.1).

The elastic net is characterised by *internal* dimension dim($G$). Every node $v_i$ in the elastic net is indexed by the discrete values of *internal coordinates* $\{\lambda_1^i,...,\lambda_{\dim(G)}^i\}$ in such a way that the nodes close on the graph have similar internal coordinates.

The purpose of the elastic net is to introduce point approximations to manifolds. Historically it was first explored and used in applications. To avoid confusion, one should notice that the term elastic net was independently introduced by several groups: for solving the traveling salesman problem (Durbin & Willshaw, 1987), in the context of principal manifolds (Gorban et al, 2001) and recently in the context of regularized regression problem (Zhou & Hastie, 2005). These three notions are completely independent and denote different things.

**Definition**. *Elastic map* is a continuous manifold $Y \in \mathbf{R}^m$ constructed from the elastic net as its grid approximation using some between-node interpolation procedure. This interpolation procedure constructs a continuous mapping $\phi_c:\{\lambda_1,..., \lambda_{\dim(G)}\} \to \mathbf{R}^m$ from the discrete map $\phi:V \to \mathbf{R}^m$, used to embed the graph in $\mathbf{R}^m$, and the discrete values of node indices $\{\lambda_1^i,...,\lambda_{\dim(G)}^i\}$, $i = 1...|V|$. For example, the simplest *piecewise linear elastic map* is built by piecewise linear map $\phi_c$.

**Definition**. *Elastic principal manifold of dimension s* for a dataset $X$ is an elastic map, constructed from an elastic net $Y$ of dimension $s$ embedded in $\mathbf{R}^m$ using such a map $\phi_{opt}:Y \to \mathbf{R}^m$ that corresponds to the minimal value of the functional

$$U^\phi(X,Y) = \mathrm{MSD}_W(X,Y) + U^\phi(G), \qquad (6)$$

where the weighted mean squared distance from the dataset $X$ to the elastic net $Y$ is calculated as the distance to the finite set of vertices $\{\mathbf{y}^1=\phi(v_1),..., \mathbf{y}^k=\phi(v_k)\}$.

In the Euclidean space one can apply an EM algorithm for estimating the elastic principal manifold for a finite dataset. It is based in turn on the general algorithm for estimating the locally optimal embedding map $\phi$ for an arbitrary elastic graph $G$, described below.

Optimisation of the elastic graph algorithm:

1) Choose some initial position of nodes of the elastic graph $\{\mathbf{y}^1=\phi(v_1),..., \mathbf{y}^k=\phi(v_k)\}$, where $k$ is the number of graph nodes $k = |V|$;
2) Calculate two matrices $e_{ij}$ and $s_{ij}$, using the following sub-algorithm:
    i. Initialize the $s_{ij}$ matrix to zero;
    ii. For each $k$-star $S_k^{(i)}$ with elasticity module $\mu_{ki}$, outer nodes $v_{N1},..., v_{Nk}$ and the central node $v_{N0}$, the $s_{ij}$ matrix is updated as follows ($1 \leq l,m \leq k$):
    $$s_{N_0 N_0} \leftarrow s_{N_0 N_0} + \mu_{ki}, \quad s_{N_l N_m} \leftarrow s_{N_l N_m} + \mu_{ki}/k^2$$
    $$s_{N_0 N_l} \leftarrow s_{N_0 N_l} - \mu_{ki}/k, \quad s_{N_l N_0} \leftarrow s_{N_l N_0} - \mu_{ki}/k$$
    iii. Initialize the $e_{ij}$ matrix to zero;
    iv. For each edge $E^{(i)}$ with weight $\lambda_i$, one vertex $v_{k1}$ and the other vertex $v_{k2}$, the $e_{jk}$ matrix is updated as follows:
    $$e_{k_1 k_1} \leftarrow e_{k_1 k_1} + \lambda_i, \quad e_{k_2 k_2} \leftarrow e_{k_2 k_2} + \lambda_i$$
    $$e_{k_1 k_2} \leftarrow e_{k_1 k_2} - \lambda_i, \quad e_{k_2 k_1} \leftarrow e_{k_2 k_1} - \lambda_i$$

3) Partition $X$ into subsets $K_i$, $i=1..k$ of data points by their proximity to $\mathbf{y}_k$: $K_i = \{\mathbf{x} : \mathbf{y}_i = \arg\min_{\mathbf{y}_j \in Y} \text{dist}(\mathbf{x}, \mathbf{y}_j)\}$;

4) Given $K_i$, calculate matrix $a_{js} = \dfrac{n_j \delta_{js}}{N \sum_{i=1}^{} w_i} + e_{js} + s_{js}$, where $n_j = \sum_{x^i \in K_j} w_i$,

   $\delta_{js}$ is the Kronecker's symbol.

5) Find new position of $\{\mathbf{y}^1,..., \mathbf{y}^k\}$ by solving the system of linear equations
$$\sum_{s=1}^{k} a_{js} \mathbf{y}^s = \dfrac{1}{N \sum_{i=1}^{} w_i} \cdot \sum_{x^i \in K_j} w_i \mathbf{x}^i$$

6) Repeat steps 3-5 until complete or approximate convergence of node positions $\{\mathbf{y}^1,..., \mathbf{y}^k\}$.

As usual, the EM algorithm described above gives only locally optimal solution. One can expect that the number of local minima of the energy function $U$ grows with increasing the 'softness' of the elastic graph (decreasing $\mu_{kj}$ parameters). Because of this, in order to obtain a solution closer to the global optimum, the *softening strategy* has been proposed, used in the algorithm for estimating the elastic principal manifold.

Algorithm for estimating the elastic principal manifold

1) Define a decreasing set of numbers $\{m_1,\ldots,m_p\}$, $m_p=1$ (for example, $\{10^3, 10^2, 10, 1\}$), defining $p$ epochs for softening;
2) Define the base values of the elastic moduli $\lambda_i^{(base)}$ and $\mu_i^{(base)}$;
3) Initialize positions of the elastic net nodes $\{\mathbf{y}^1,\ldots,\mathbf{y}^k\}$ on the linear principal manifold spanned by first $\dim(G)$ principal components;
4) Set *epoch_counter* = 1
5) Set the elastic moduli $\lambda_i = m_{epoch\_counter}\lambda_i^{(base)}$ and $\mu_i = m_{epoch\_counter}\mu_i^{(base)}$;
6) Modify the elastic net using the algorithm for optimisation of the elastic graph;
7) Repeat steps 5-6 for all values of *epoch_counter* = 2, ... , $p$.

**Remark.** The values $\lambda_i$ and $\mu_j$ are the coefficients of stretching elasticity of every edge $E^{(i)}$ and of bending elasticity of every rib $S_2^{(j)}$. In the simplest case $\lambda_1 = \lambda_2 = ... = \lambda_s = \lambda(s)$, $\mu_1 = \mu_2 = ... = \mu_r = \mu(r)$, where $s$ and $r$ are the numbers of edges and ribs correspondingly. Approximately dependence on graph 'resolution' is given by Gorban & Zinovyev (2007): $\lambda(s) = \lambda_0 \cdot s^{\frac{2-\dim(G)}{\dim(G)}}$, $\mu(s) = \mu_0 \cdot r^{\frac{2-\dim(G)}{\dim(G)}}$. This formula is applicable, of course, only for the elastic nets. In general a case $\lambda_i$ and $\mu_i$ are often made variable in different parts of the graph accordingly to some adaptation strategy (Gorban & Zinovyev, 2005).

**Remark.** $U_E^\phi(G)$ penalizes the total length (or, indirectly, 'square', 'volume, etc.) of the constructed manifold and provides regularization of distances between node positions at the initial steps of the softening. At the final stage of the softening $\lambda_i$ can be put to zero with little effect on the manifold configuration.

*Elastic map post-processing* such as *map extrapolation* can be applied to increase its usability and avoid the 'border effect', for details see (Gorban & Zinovyev, 2008a).

PLURIHARMONIC GRAPHS AS IDEAL APPROXIMATORS

Approximating datasets by one dimensional principal curves is not satisfactory in the case of datasets that can be intuitively characterized as *branched*. A principal object which naturally passes through the 'middle' of such a data distribution should also have branching points that are missing in the simple structure of principal curves. Introducing such branching points converts principal curves into *principal graphs*.

Principal graphs were introduced by Kégl & Krzyzak (2002) as a natural extension of one-dimensional principal curves in the context of skeletonisation of hand-written symbols. The most important part of this definition is the form of the penalty imposed onto deviation of the configuration of the branching points embedment from their 'ideal' configuration (*end*, *line*, *corner*, *T-*, *Y-* and *X-*configuration). Assigning types for all vertices serves for definition of the penalty on the total deviation from the graph 'ideal' configuration (Kégl, 1999). Other types of vertices were not considered, and outside the field of symbol skeletonization applicability of such a definition of principal graph remains limited.

Gorban & Zinovyev (2005), Gorban et al. (2007), and Gorban et al. (2008) proposed to use a universal form of non-linearity penalty for the branching points. The form of this penalty is defined in the previous chapter for the elastic energy of graph embedment. It naturally generalizes the simplest three-point *second derivative* approximation squared:

for a 2-star (or rib) the penalty equals $\| \phi(S_2^{(j)}(0)) - \frac{1}{2}(\phi(S_2^{(j)}(1)) + \phi(S_2^{(j)}(2))) \|^2$,

for a 3-star it is $\| \phi(S_3^{(j)}(0)) - \frac{1}{3}(\phi(S_3^{(j)}(1)) + \phi(S_3^{(j)}(2)) + \phi(S_3^{(j)}(3))) \|^2$, etc.

For a *k*-star this penalty equals to zero iff the position of the central node coincides with the mean point of its neighbors. An embedment $\phi(G)$ is 'ideal' if all such penalties equal to zero. For a *primitive elastic graph* this means that this embedment is a *harmonic function on graph*: its value in each non-terminal vertex is a mean of the value in the closest neighbors of this vertex.

For non-primitive graphs we can consider stars which include not all neighbors of their centers. For example, for a square lattice we create elastic graph (elastic net) using 2-stars (ribs): all vertical 2-stars and all horizontal 2-stars. For such elastic net, each non-boundary vertex belongs to two stars. For a general elastic graph *G* with sets of *k*-stars $S_k$ we introduce the following notion of pluriharmoning function.

**Definition.** A map $\phi: V \to \mathbf{R}^m$ defined on vertices of *G* is *pluriharmonic* iff for any *k*-star $S_k^{(j)} \in S_k$ with the central vertex $S_k^{(j)}(0)$ and the neighbouring vertices $S_k^{(j)}(i)$, $i = 1...k$, the equality holds:

$$\phi(S_k^{(j)}(0)) = \frac{1}{k} \sum_{i=1}^{k} \phi(S_k^{(j)}(i)). \tag{7}$$

Pluriharmonic maps generalize the notion of linear map and of harmonic map, simultaneously. For example:

1) 1D harmonic functions are linear;
2) If we consider an *n*D cubic lattice as a primitive graph (with 2*n*-stars for all non-boundary vertices), then the correspondent pluriharmonic functions are just harmonic ones;
3) If we create from *n*D cubic lattice a standard *n*D elastic net with 2-stars (each non-boundary vertex is a center of *n* 2-stars, one 2-stars for each coordinate direction), then pluriharmonic functions are linear.

Pluriharmonic functions have many attractive properties, for example, they satisfy the following *maximum principle*. A vertex *v* of an elastic graph is called a *corner point* or an *extreme point* of *G* iff *v* is not a centre of any *k*-star from $S_k$ for all *k*>0.

**Theorem.** Let $\phi:V\to \mathbf{R}^m$ be a pluriharmonic map, $F$ be a convex function on $\mathbf{R}^m$, and $a = \max_{x\in V}F(\phi(x))$. Then there is a corner point $v$ of $G$ such that $F(\phi(v))=a$.

Convex functions achieve their maxima in corner points. Even a particular case of this theorem with linear functions $F$ is quite useful. Linear functions achieve their maxima and minima in corner points.

In the theory of principal curves and manifolds the penalty functions were introduced to penalise deviation from linear manifolds (straight lines or planes). We proposed to use pluriharmonic embeddings ('pluriharmonic graphs') as 'ideal objects' instead of manifolds and to introduce penalty (5) for deviation from this ideal form.

GRAPH GRAMMARS AND THREE TYPES OF COMPLEXITY FOR PRINCIPAL GRAPHS

Principal graphs can be called *data approximators of controllable complexity*. By complexity of the principal objects we mean the following three notions:

1) *Geometric complexity*: how far a principal object deviates from its ideal configuration; for the elastic principal graphs we explicitly measure deviation from the 'ideal' pluriharmonic graph by the elastic energy $U_\phi(G)$ (3) (this complexity may be considered as a *measure of non-linearity*);
2) *Structural complexity measure*: it is some non-decreasing function of the number of vertices, edges and $k$-stars of different orders $SC(G)=SC(|V|,|E|,|S_2|,\ldots,|S_m|)$; this function penalises for number of structural elements;
3) *Construction complexity* is defined with respect to a graph grammar as a number of applications of elementary transformations necessary to construct given $G$ from the simplest graph (one vertex, zero edges).

The construction complexity is defined with respect to a *grammar* of elementary transformation. The graph grammars (Löwe, 1993; Nagl, 1976) provide a well-developed formalism for the description of elementary transformations. An elastic graph grammar is presented as a set of production (or substitution) rules. Each rule has a form $A \to B$, where $A$ and $B$ are elastic graphs. When this rule is applied to an elastic graph, a copy of $A$ is removed from the graph together with all its incident edges and is replaced with a copy of $B$ with edges that connect $B$ to the graph. For a full description of this language we need the notion of a *labeled graph*. Labels are necessary to provide the proper connection between $B$ and the graph (Nagl, 1976). An approach based on *graph grammars* to constructing effective approximations of an *elastic principal graph* has been recently proposed (Gorban et al, 2007).

Let us define *graph grammar O* as a set of graph grammar operations $O=\{o_1,..,o_s\}$. All possible applications of a graph grammar operation $o_i$ to a graph $G$ gives a set of transformations of the initial graph $o_i(G) = \{G_1, G_2, \ldots, G_p\}$, where $p$ is the number of all possible applications of $o_i$ to $G$. Let us also define a sequence of $r$ different graph grammars
$\{O^{(1)} = \{o_1^{(1)},...,o_{s_1}^{(1)}\}, \cdots, O^{(r)} = \{o_1^{(r)},...,o_{s_r}^{(r)}\}\}$.

Let us choose a grammar of elementary transformations, predefined boundaries of structural complexity $SC_{max}$ and construction complexity $CC_{max}$, and elasticity coefficients $\lambda_i$ and $\mu_{kj}$.

**Definition**. *Elastic principal graph* for a dataset $X$ is such an elastic graph $G$ embedded in the Euclidean space by the map $\phi: V \to \mathbf{R}^m$ that $SC(G) \leq SC_{max}$, $CC(G) \leq CC_{max}$, and $U_\phi(G) \to \min$ over all possible elastic graphs $G$ embeddings in $\mathbf{R}^m$.

Algorithm for estimating the elastic principal graph

1) Initialize the elastic graph $G$ by 2 vertices $v_1$ and $v_2$ connected by an edge. The initial map $\phi$ is chosen in such a way that $\phi(v_1)$ and $\phi(v_2)$ belong to the first principal line in such a way that all the data points are projected onto the principal line segment defined by $\phi(v_1)$, $\phi(v_2)$;
2) For all $j=1\ldots r$ repeat steps 3-6:
3) Apply all grammar operations from $O^{(j)}$ to $G$ in all possible ways; this gives a collection of candidate graph transformations $\{G_1, G_2, \ldots\}$;
4) Separate $\{G_1, G_2, \ldots\}$ into *permissible* and *forbidden* transformations; permissible transformation $G_k$ is such that $SC(G_k) \leq SC_{max}$, where $SC_{max}$ is some predefined structural complexity ceiling;
5) Optimize the embedment $\phi$ and calculate the elastic energy $U_\phi(G)$ of graph embedment for every permissible candidate transformation, and choose such a graph $G_{opt}$ that gives the minimal value of the elastic functional:
$$G_{opt} = \arg\inf_{G_k \in permissible\ set} U_\phi(G_k);$$
6) Substitute $G \to G_{opt}$;
7) Repeat steps 2-6 until the set of permissible transformations is empty or the number of operations exceeds a predefined number – the construction complexity.

PRINCIPAL TREES AND METRO MAPS

Let us construct the simplest non-trivial type of the principal graphs, called principal trees. For this purpose let us introduce a simple 'Add a node, bisect an edge' graph grammar (see Fig. 2) applied for the class of primitive elastic graphs.

**Definition**. *Principal tree* is an acyclic primitive elastic principal graph.

**Definition**. *'Remove a leaf, remove an edge' graph grammar* $O^{(shrink)}$ applicable for the class of primitive elastic graphs consists of two operations: 1) The transformation *'remove a leaf'* can be applied to any vertex $v$ of $G$ with connectivity degree equal to 1: remove $v$ and remove the edge $(v,v')$ connecting v to the tree; 2) The transformation *'remove an edge'* is applicable to any pair of graph vertices $v, v'$ connected by an edge $(v, v')$: delete edge $(v, v')$, delete vertex $v'$, merge the $k$-stars for which $v$ and $v'$ are the central nodes and make a new $k$-star for which $v$ is the central node with a set of neighbors which is the union of the neighbors from the $k$-stars of $v$ and $v'$.

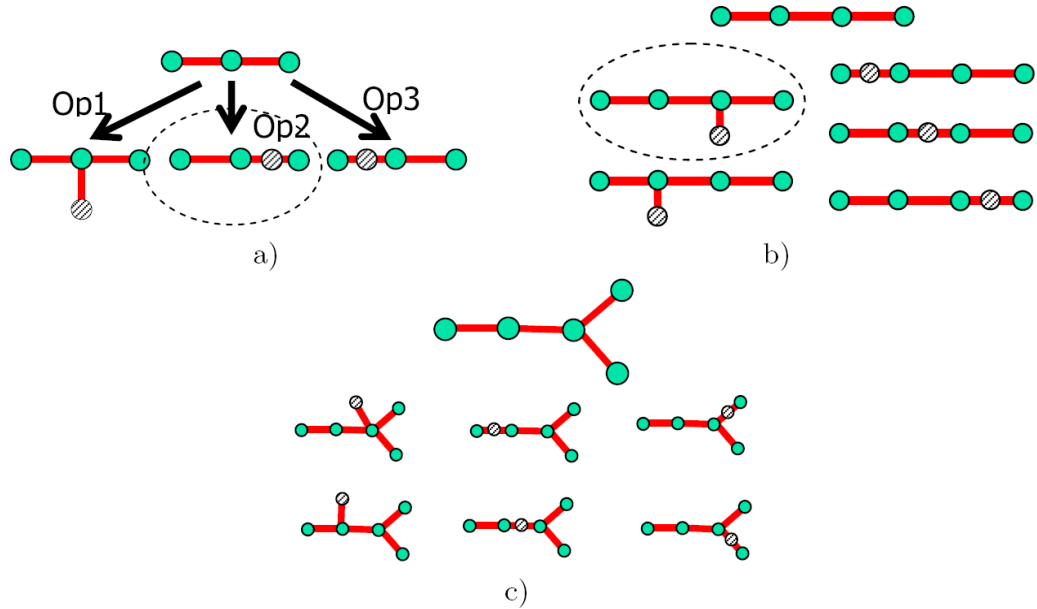

*Fig. 2. Illustration of the simple "add node to a node" or "bisect an edge" graph grammar. a) We start with a simple 2-star from which one can generate three distinct graphs shown. The "Op1" operation is adding a node to a node, operations "Op1" and "Op2" are edge bisections (here they are topologically equivalent to adding a node to a terminal node of the initial 2-star). For illustration let us suppose that the "Op2" operation gives the biggest elastic energy decrement, thus it is the "optimal" operation. b) From the graph obtained one can generate 5 distinct graphs and choose the optimal one. c) The process is continued until a definite number of nodes are inserted.*

**Definition**. *'Add a node, bisect an edge' graph grammar* $O^{(grow)}$ applicable for the class of primitive elastic graphs consists of two operations: 1) The transformation *"add a node"* can be applied to any vertex $v$ of $G$: add a new node $z$ and a new edge $(v, z)$; 2) The transformation *"bisect an edge"* is applicable to any pair of graph vertices $v, v'$ connected by an edge $(v, v')$: delete edge $(v, v')$, add a vertex $z$ and two edges, $(v, z)$ and $(z, v')$. The transformation of the elastic structure (change in the star list) is induced by the change of topology, because the elastic graph is primitive. Consecutive application of the operations from this grammar generates trees, i.e. graphs without cycles.

Also we should define the structural complexity measure $SC(G)=SC(|V|,|E|,|S_2|,\ldots,|S_m|)$. Its concrete form depends on the application field. Here are some simple examples:

1) $SC(G) = |V|$ : i.e., the graph is considered more complex if it has more vertices;

2) $SC(G) = \begin{cases} |S_3|, & \text{if } |S_3| \leq b_{max} \text{ and } \sum_{k=4}^{m}|S_k| = 0 \\ \infty, & \text{otherwise} \end{cases}$,

   i.e., only $b_{max}$ simple branches (3-stars) are allowed in the principal tree.

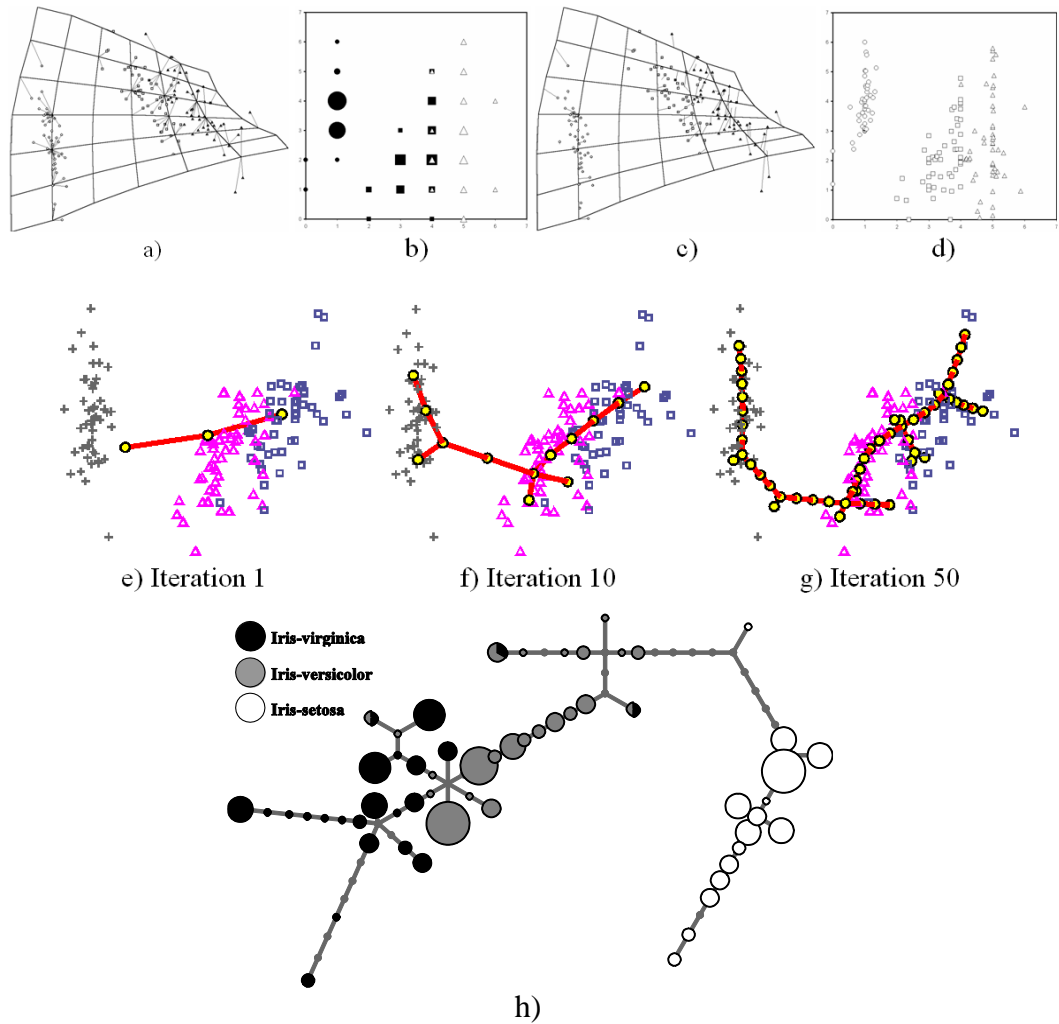

*Fig. 3. Principal manifold and principal tree for the Iris dataset. a) View of the principal manifold projected on the first two principal components, the data points are shown projected into the closest vertex of the elastic net; b) visualization of data points in the internal coordinates, here classes are represented in the form of Hinton diagrams: the size of the diagram is proportional to the number of points projected, the shape of the diagram denote three different point classes; c) same as a), but the data points are shown projected into the closest point of the piecewise linearly interpolated elastic map; d) same as b), but based on projection shown in c); e)-g) First 50 iterations of the principal tree algorithm, the tree is shown projected onto the principal plane; h) metro map representation of the Iris dataset.*

Using the sequence $\{O^{(grow)}, O^{(grow)}, O^{(shrink)}\}$ in the above-described algorithm for estimating the elastic principal graph gives an approximation to the principal trees. Introducing the 'tree trimming' grammar $O^{(shrink)}$ allows to produce principal trees closer to the global optimum, trimming excessive tree branching and fusing $k$-stars separated by small 'bridges'.

Principal trees can have applications in data visualization. A principal tree is embedded into a multidimensional data space. It approximates the data so that one can project points from the multidimensional space into the closest node of the tree. The tree by its construction is a one-dimensional object, so this projection performs dimension reduction of the multidimensional data. The question is how to produce a planar tree layout? Of course, there are many ways to layout a tree on a plane without edge intersection. But it would be useful if both local tree properties and global distance relations would be represented using the layout. We can require that

1) In a two-dimensional layout, all *k*-stars should be represented equiangular; this is the small penalty configuration;
2) The edge lengths should be proportional to their length in the multidimensional embedding; thus one can represent between-node distances.

This defines a tree layout up to global rotation and scaling and also up to changing the order of leaves in every *k*-star. We can change this order to eliminate edge intersections, but the result can not be guaranteed. In order to represent the global distance structure, it was found (Gorban et al., 2008) that a good approximation for the order of *k*-star leaves can be taken from the projection of every *k*-star on the linear principal plane calculated for all data points, or on the local principal plane in the vicinity of the *k*-star, calculated only for the points close to this star. The resulting layout can be further optimized using some greedy optimization methods.

The point projections are then represented as pie diagrams, where the size of the diagram reflects the number of points projected into the corresponding tree node. The sectors of the diagram allow us to show proportions of points of different classes projected into the node (see an example on Fig. 3).

This data display was called a *"metro map"* since it is a schematic and "idealized" representation of the tree and the data distribution with inevitable distortions made to produce a 2D layout. However, using this map one can still estimate the distance from a point (tree node) to a point passing through other points. This map is inherently unrooted (as a real metro map). It is useful to compare this metaphor with trees produced by hierarchical clustering where the metaphor is closer to a "genealogy tree".

PRINCIPAL CUBIC COMPLEXES

Elastic nets introduced above are characterized by their internal dimension $\dim(G)$. The way to generalize these characteristics on other elastic graphs is to utilize the notion of cubic complex (Gorban et al, 2007).

**Definition**. *Elastic cubic complex K of internal dimension r* is a Cartesian product $G_1 \times ... \times G_r$ of elastic graphs $G_1, ... G_r$. It has the vertex set $V_1 \times ... \times V_r$. Let $1 \le i \le r$ and $v_j \in V_j$ ($j \ne i$). For this set of vertices, $\{v_j\}_{j \ne i}$, a copy of $G_i$ in $G_1 \times ... \times G_r$ is defined with vertices $(v_1, ..., v_{i-1}, v, v_{i+1}, ..., v_r)$ ($v \in V_i$), edges

$$((v_1, ..., v_{i-1}, v, v_{i+1}, ..., v_r), (v_1, ..., v_{i-1}, v', v_{i+1}, ..., v_r)), (v, v') \in E_i,$$

and, similarly, $k$-stars of the form $(v_1, ..., v_{i-1}, S_k, v_{i+1}, ..., v_r)$, where $S_k$ is a $k$-star in $G_i$. For any $G_i$ there are $\prod_{j \neq i} |V_j|$ copies of $G_i$ in $G$. Sets of edges and $k$-stars for Cartesian product are unions of that set through all copies of all factors. A map $\varphi : V_1 \times ... \times V_r \to \mathbf{R}^m$ maps all the copies of factors into $\mathbf{R}^m$ too.

**Remark.** By construction, *the energy of the elastic graph product is the energy sum of all factor copies.* It is, of course, a quadratic functional of $\phi$.

If we approximate multidimensional data by an $r$-dimensional object, the number of points (or, more generally, elements) in this object grows with $r$ exponentially. This is an obstacle for grammar–based algorithms even for modest $r$, because for analysis of the rule $A \to B$ applications we should investigate all isomorphic copies of $A$ in $G$. Introduction of a cubic complex is useful factorization of the principal object which allows to avoid this problem.

The only difference between the construction of general elastic graphs and factorized graphs is in the application of the transformations. For factorized graphs, we apply them to factors. This approach significantly reduces the amount of trials in selection of the optimal application. The simple grammar with two rules, "add a node to a node, or bisect an edge," is also powerful here, it produces products of primitive elastic trees. For such a product, the elastic structure is defined by the topology of the factors.

INCOMPLETE DATA

Some of the methods described above allow us to use incomplete data in a natural way. Let us represent an incomplete observation by $\mathbf{x} = (x_1, ..., @, ..., @, ..., x_m)$, where the '@' symbol denotes a missing value.

**Definition.** *Scalar product between two incomplete observations* $\mathbf{x}$ *and* $\mathbf{y}$ *is* $(\mathbf{x}, \mathbf{y}) = \sum_{i \neq @}^m x_i y_i$. Then the Euclidean distance is $\sqrt{(\mathbf{x} - \mathbf{y})^2} = \sqrt{\sum_{i \neq @}^m (x_i - y_i)^2}$.

**Remark.** This definition has a very natural geometrical interpretation: an incomplete observation with $k$ missing values is represented by a $k$–dimensional linear manifold $L_k$, parallel to $k$ coordinate axes corresponding to the missing data.

Thus, any method which uses only scalar products or/and Euclidean distances can be applied for incomplete data with some minimal modifications subject to random and not too dense distribution of missing values in $X$. For example, the iterative method for SVD for incomplete data matrix was developed (Roweis (1998); Gorban & Rossiev, 1999).

There are, of course, other approaches to incomplete data in unsupervised learning (for example, those presented by Little & Rubin (1987)).

IMPLICIT METHODS

Most of the principal objects introduced in this paper are constructed as explicit geometrical objects embedded in $\mathbf{R}^m$ to which we can calculate the distance from any object in $X$. In this way, they generalize the "data approximation"-based (#1) and the "variation-maximization"-based (#2) definitions of linear PCA. There also exists the whole family of methods, which we only briefly mention here, that generalize the "distance distortion minimization" definition of PCA (#3).

First, some methods take as input a pairwise distance (or, more generally, *dissimilarity*) matrix $D$ and construct such a configuration of points in a low-dimensional Euclidean space that the distance matrix $D'$ in this space reproduce $D$ with maximal precision. The most fundamental in this series is the *metric multidimensional scaling* (Kruskal, 1964). The next is the Kernel PCA approach (Schölkopf et al., 1997) which takes advantage of the fact that for the linear PCA algorithm one needs only the matrix of pairwise scalar products (Gramm matrix) but not the explicit values of coordinates of $X$. It allows to apply the kernel trick (Aizerman et al., 1964) and substitute the Gramm matrix by the scalar products calculated with use of some kernel functions. Kernel PCA method is tightly related to the classical multidimensional scaling (Williams, 2002).

Local Linear Embedding or LLE (Roweis & Saul, 2000) searches for such a $N \times N$ matrix $A$ that approximates given $\mathbf{x}^i$ by a linear combination of $n$ vectors-neighbours of $\mathbf{x}^i$: $\sum_{i=1}^{N} \| \mathbf{x}^i - \sum_{k=1}^{N} A_k^i \mathbf{x}^k \|^2 \to \min$, where only such $A_k^i \neq 0$, if $k$ is one of the $n$ closest to $\mathbf{x}^i$ vectors. After one constructs such a configuration of points in $\mathbf{R}^s$, $s \ll m$, that $\mathbf{y}^i = \sum_{k=1}^{N} A_k^i \mathbf{y}^k$, $\mathbf{y}^i \in \mathbf{R}^s$, for all $i = 1 \ldots N$. The coordinates of such embedding are given by the eigenvectors of the matrix $(1-A)^T(1-A)$.

ISOMAP (Tenenbaum et al., 2000) and Laplacian eigenmap (Belkin & Niyogi, 2003; Nadler et al., 2008) methods start with construction of the *neighbourhood graph*, i.e. the graph in which close in some sense data points are connected by (weighted) edges. This weighted graph can be represented in the form of a weighted *adjacency matrix* $W = \{W_{ij}\}$. From this graph, ISOMAP constructs a new distance matrix $D^{(ISOMAP)}$, based on the path lengths between two points in the neighbourhood graph, and the multidimensional scaling is applied to $D^{(ISOMAP)}$. The Laplacian map solves the eigenproblem $L\mathbf{f}_\lambda = \lambda S \mathbf{f}_\lambda$, where $S = diag\{\sum_{j=1}^{N} W_{0j}, \cdots, \sum_{j=1}^{N} W_{Nj}\}$, $L = S - W$ is the Laplacian matrix. The trivial constant solution corresponding to the smallest eigenvalue $\lambda_0 = 0$ is discarded, while the elements of the eigenvectors $\mathbf{f}_{\lambda_1}, \mathbf{f}_{\lambda_2}, \cdots, \mathbf{f}_{\lambda_s}$, where $\lambda_1 < \lambda_2 < \ldots < \lambda_s$, give the $s$-dimensional projection of $\mathbf{x}^i$, i.e. $P(\mathbf{x}^i) = \{\mathbf{f}_{\lambda_1}(i), \mathbf{f}_{\lambda_2}(i), \cdots, \mathbf{f}_{\lambda_s}(i)\}$.

Finally, one can implicitly construct projections into smaller dimensional spaces by training *auto-associative neural networks* with narrow *hidden layer*. An overview of the existing Neural PCA methods can be found in the recent collection of review papers (Gorban et al, 2008).

EXAMPLE: PRINCIPAL OBJECTS FOR THE IRIS DATASET

On Fig. 3 we show application of the elastic principal manifolds and principal trees algorithms to the standard Iris dataset (Fisher, 1936). As expected, two-dimensional approximation of the principal manifold in this case is close to the linear principal plane. One can also see that the principal tree illustrates well the fact of almost complete separation of classes in data space.

EXAMPLE: PRINCIPAL OBJECTS FOR MOLECULAR SURFACES

A molecular surface defines the effective region of space which is occupied by a molecule. For example, the Van-der-Waals molecular surface is formed by surrounding every atom in the molecule by a sphere of radius equal to the characteristic radius of the Van-der-Waals force. After all the interior points are eliminated, this forms a complicated non-smooth surface in 3D. In practice, this surface is sampled by a finite number of points.

Using principal manifolds methodology, we constructed a smooth approximation of such molecular surface for a small piece of a DNA molecule (several nucleotides long). First, we have made an approximation of this dataset by a 1D principal curve. Interestingly, this curve followed the backbone of the molecule, forming a helix (see Fig. 4). Second, we approximated the molecular surface by a 2D manifold. The topology of the surface is expected to be spherical, so we applied spherical topology of the elastic net for optimisation.

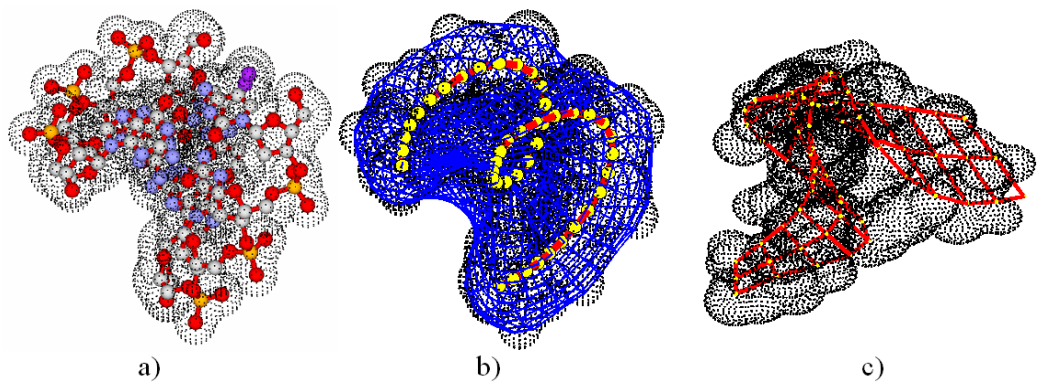

a)   b)   c)

*Fig. 4. Principal objects approximating molecular surface of a short stretch of DNA molecule. a) stick-and-balls model of the DNA stretch and the initial molecular surface (black points); b) one- and two-dimensional spherical principal manifolds for the molecular surface; c) simple principal cubic complex (product of principal trees) which does not have any branching in this case.*

We should notice that since it is impossible to make the lengths of all edges equal for the spherical grid, corrections were performed for the edge elasticities during the grid initialization (shorter edges are given larger $\lambda_i$s). Third, we applied the method for constructing principal cubic complexes, namely, graph product of principal trees, which produced somewhat trivial construction (because no branching was energetically optimal): product of two short elastic principal curves, forming a double helix.

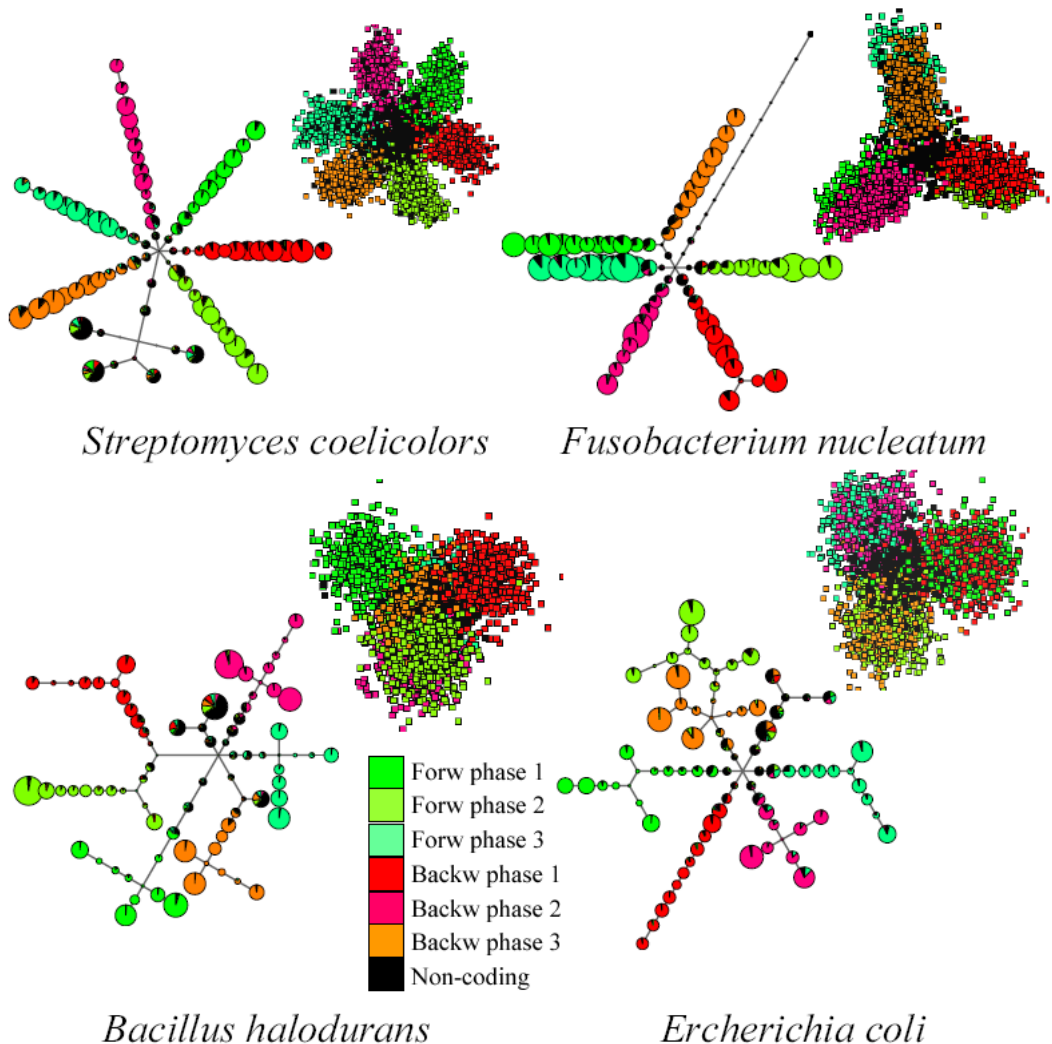

Fig. 5. Seven cluster structures presented for 4 selected genomes. A genome is represented as a collection of points (text fragments represented by their triplet frequencies) in the 64-multidimensional space. Color codes denote point classes corresponding to 6 possible frameshifts when a random fragment overlaps with a coding gene (3 in the forward and 3 in the backward direction of the gene), and the black color corresponds to non-coding regions. For every genome a principal tree ("metro map" layout) is shown together with 2D PCA projection of the data distribution. Note that the clusters that appear to be mixed on the PCA plot for Escherichia coli (they remain mixed in 3D PCA as well) are well separated on the "metro map". This proves that they are well-separated in $\mathbf{R}^{64}$.

EXAMPLE: PRINCIPAL OBJECTS DECIPHER GENOME

A dataset *X* can be constructed for a string sequence using a short word frequency dictionary approach in the following way: 1) the notion of word is defined; 2) the set of all possible short words is defined, let us say that we have *m* of them; 3) a number *N* of text fragments of certain width is sampled from the text; 4) in each fragment the frequency of occurrences of all possible short words is calculated and, thus, each fragment is represented as a vector in multidimensional space $\mathbf{R}^m$. The whole text then is represented as a dataset of $N$ vectors embedded in $\mathbf{R}^m$.

We systematically applied this approach to available bacterial genomic sequences (Gorban & Zinovyev, 2008b). In our case we defined: 1) a word is a sequence of three letters from the {$A,C,G,T$} alphabet (triplet); 2) evidently, there are 64 possible triplets in the {$A,C,G,T$} alphabet; 3) we sampled 5000-10000 fragments of width 300 from a genomic sequence; 4) we calculated the frequencies of non-overlapping triplets for every fragment.

The constructed datasets are interesting objects for data-mining, because 1) they have a non-trivial cluster structure which usually contains various configurations of 7 clusters (see Fig. 5); 2) class labels can be assigned to points accordingly to available genome annotations; in our case we put information about presence (in one of six possible frameshifts) or absence of the coding information in the current position of a genome; 3) using data mining techniques here has immediate applications in the field of automatic gene recognition and in others, see, for example, (Carbone et al, 2003). On Fig. 5 we show application of both classical PCA and the metro map methods for several bacterial genomes. Look at http://www.ihes.fr/~zinovyev/7clusters web-site for further information.

EXAMPLE: NON-LINEAR PRINCIPAL MANIFOLDS FOR MICROARRAY DATA VISUALIZATION

DNA microarray data is a rich source of information for molecular biology (an expository overview is provided by Leung & Cavalieri (2003)). This technology found numerous applications in understanding various biological processes including cancer. It allows to screen simultaneously the expression of all genes in a cell exposed to some specific conditions (for example, stress, cancer, treatment, normal conditions). Obtaining a sufficient number of observations (chips), one can construct a table of "samples vs genes", containing logarithms of the expression levels of, typically several thousands ($n$) of genes, in typically several tens ($m$) of samples.

On Fig. 6 we provide a comparison of data visualization scatters after projection of the breast cancer dataset, provided by Wang et al. (2003), onto the linear two- and non-linear two-dimensional principal manifold. The latter one is constructed by the elastic maps approach. Each point here represents a patient treated from cancer. Before dimension reduction it is represented as a vector in $\mathbf{R}^n$, containing the expression values for all $n$ genes in the tumor sample. Linear and non-linear 2D principal manifolds provide mappings $\mathbf{R}^n \to \mathbf{R}^2$, drastically reducing vector dimensions and allowing data visualization. The form, the shape and the size of the point on the Fig.6 represent various clinical data (class labels) extracted from the patient's disease records.

Practical experience from bioinformatics studies shows that two-dimensional data visualization using non-linear projections allow to catch more signals from data (in the form of clusters or specific regions of higher point density) than linear projections, see Fig. 6 and a good example by Ivakhno & Armstrong (2008).

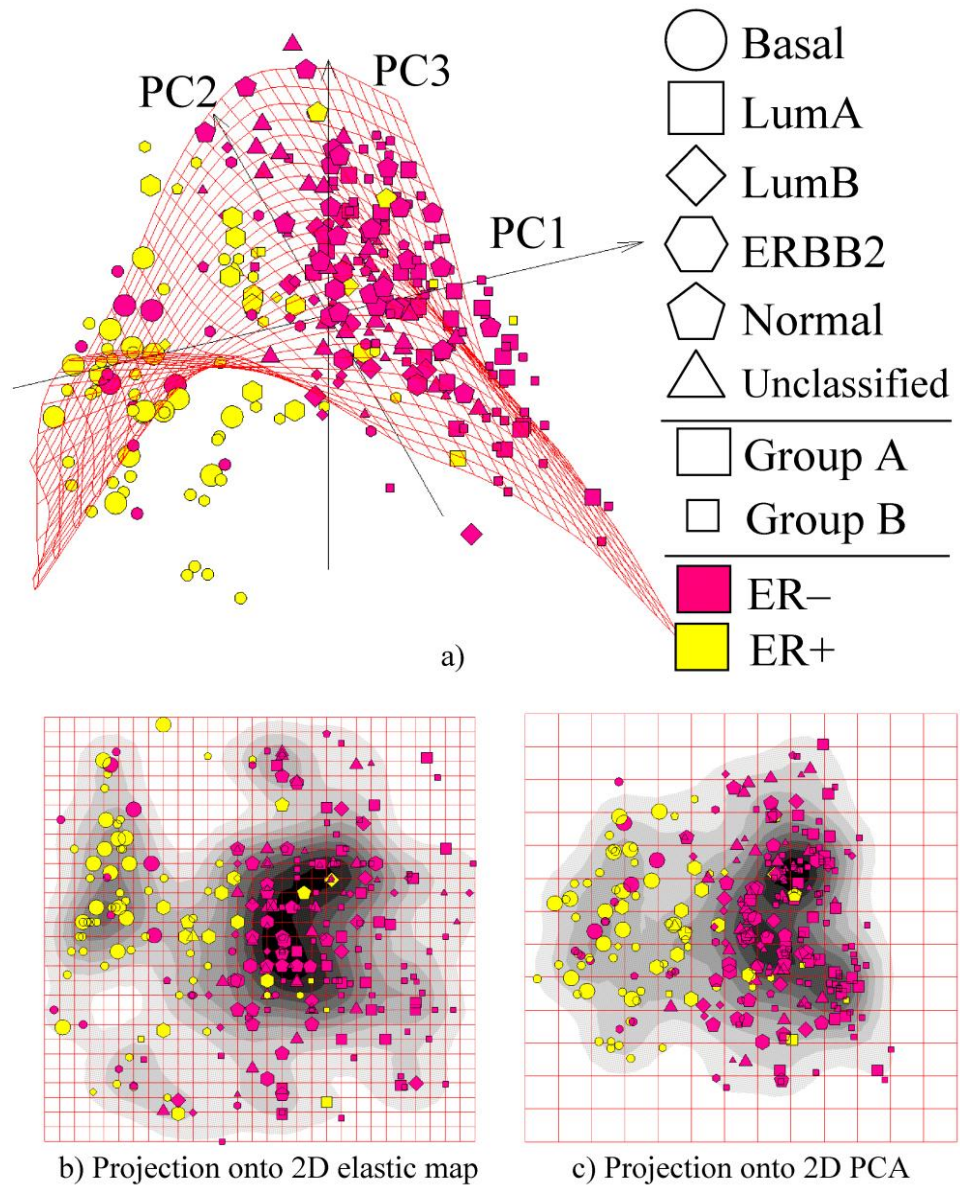

*Figure 6. Visualization of breast cancer microarray dataset using elastic maps. Ab initio classifications are shown using points size (ER, estrogen receptor status), shape (Group A – patients with aggressive cancer, Group B – patients with non-aggressive cancer) and color (TYPE, molecular type of breast cancer). a) Configuration of nodes projected into the three-dimensional principal linear manifold. One clear feature is that the dataset is curved such that it can not be mapped adequately onto a two-dimensional principal plane. b) The distribution of points in the internal non-linear manifold coordinates is shown together with estimation of the two-dimensional density of points. c) The same as b) but for the linear two-dimensional manifold. One can notice that the ``basal'' breast cancer subtype is much better separated on the non-linear mapping and some features of the distribution become better resolved.*

In addition to that, Gorban & Zinovyev (2008a) performed a systematic comparison of performance of low-dimensional linear and non-linear principal manifolds for microarray data visualization, using the following four criteria: 1) mean-square distance error; 2) distortions in mapping the big distances between points; 3) local point neighbourhood preservation; 4) compactness of point class labels after projection. It was demonstrated that non-linear two-dimensional principal manifolds provide systematically better results accordingly to all these criteria, achieving the performance of three- and four- dimensional linear principal manifolds (principal components).

The interactive ViMiDa (Visualization of Microarray Data) and ViDaExpert software allowing microarray data visualization with use of non-linear principal manifolds are available on the web-site of Institut Curie (Paris): *http://bioinfo.curie.fr/projects/vidaexpert* and *http://bioinfo.curie.fr/projects/vimida*.

CONCLUSION

In this chapter we gave a brief practical introduction into the methods of construction of principal objects, i.e. objects embedded in the 'middle' of the multidimensional data set. As a basis, we took the unifying framework of mean squared distance approximation of finite datasets which allowed us to look at the principal graphs and manifolds as generalizations of the mean point notion.

BIOGRAPHY

Prof. Dr. Alexander Gorban obtained his PhD degree in differential equations and mathematical physics in 1980, and Dr. Sc. degree in biophysics in 1990. He holds now a chair of Applied Mathematics at the University of Leicester, UK, and he is the Chief Scientist at the Institute for Computational Modelling Russian Academy of Sciences (Krasnoyarsk, Russia). His scientific interest include interdisciplinary problem of model reduction, topological dynamics, physical and chemical kinetics, mathematical biology and data mining.

Dr. Andrei Zinovyev obtained his university formation in theoretical physics (cosmology). In 2001 he obtained his PhD degree in Computer Science at the Institute for Computational Modelling of Russian Academy of Sciences (Krasnoyarsk, Russia). After defending his PhD, he moved to France, at Institut des Hautes Etudes Scientifiques in Bures-sur-Yvette to work as a post-doc during 3 years in the mathematical biology group of Professor Misha Gromov. Since January 2005, he is the head of the computational systems biology of cancer team at Institut Curie (INSERM Unit U900 "Bioinformatics and Computational Systems Biology of Cancer") in Paris. His main area of scientific expertise is bioinformatics, systems biology of cancer, dimension reduction in high-throughput data analysis and model reduction in the dynamical models.

KEY TERMS AND THEIR DEFINITIONS

**Principal components:** such an orthonormal basis in which the covariance matrix is diagonal.

**Principal manifold:** intuitively, a smooth manifold going through the middle of data cloud; formally, there exist several definitions for the case of data distributions: 1) Hastie and Stuelze's principal manifolds are self-consistent curves and surfaces; 2) Kegl's principal curves provide the minimal mean squared error given the limited curve length; 3) Tibshirani's principal curves maximize the likelihood of the additive noise data model; 4) Gorban and Zinovyev elastic principal manifolds minimize a mean square error functional regularized by addition of energy of manifold stretching and bending; 5) Smola's regularized principal manifolds minimize some form of a regularized quantization error functional; and some other definitions.

**Principal graph:** a graph embedded in the multidimensional data space, providing the minimal mean squared distance to the dataset combined with deviation from an "ideal" configuration (for example, from pluriharmonic graph) and not exceeding some limits on complexity (in terms of the number of structural elements and the number of graph grammar transformations needed for obtaining the principal graph from some minimal graph).

**Self-consistent approximation:** approximation of a dataset by a set of vectors such that every point **y** in the vector set is a conditional mean of all points from dataset that are projected in **y**.

**Expectation/Maximisation algorithm:** generic splitting algorithmic scheme with use of which almost all algorithms for estimating principal objects are constructed; it consists of two basic steps: 1) projection step, at which the data is projected onto the approximator, and 2) maximization step, at which the approximator is optimized given the projections obtained at the previous step.

# Principal Graphs and Manifolds

Alexander Gorban
Department of Mathematics, University of Leicester
University Road, Leicester LE1 7RH, United Kingdom
+44 (0) 116 223 14 33
ag153@le.ac.uk

Andrei Zinovyev
Institut Curie
26 rue d'Ulm, Paris 75248
+33 (0) 56 24 69 89
andrei.zinovyev@curie.fr